\documentclass[]{article}
\usepackage{geometry}
\usepackage[backend=biber]{biblatex}
\usepackage{authblk}
\usepackage{graphicx}
\usepackage{epstopdf}
\usepackage{multirow}
\usepackage{booktabs}
\usepackage{amsmath}
\usepackage{amssymb}
\usepackage{bbm}
\usepackage{lmodern}
\usepackage{titlesec}
\usepackage{hyperref}
\usepackage{dblfloatfix}
\usepackage{subcaption}

\titleformat{\subsubsection}[runin]
  {\normalfont\normalsize\bfseries}{\thesubsubsection}{1em}{}
\titlespacing*{\subsubsection}{0pt}{0.7ex}{1.5ex plus .2ex}

\title{DeepPhys: Video-Based Physiological Measurement Using Convolutional Attention Networks}
\author[1]{Weixuan Chen}
\author[2]{Daniel McDuff}

\affil[1]{Massachusetts Institute of Technology, Cambridge, Massachusetts}
\affil[2]{Microsoft Research, Redmond, Washington}

\date{March 2018}

\begin{document}


\maketitle

\begin{abstract}
Non-contact video-based physiological measurement has many applications in health care and human-computer interaction. Practical applications require measurements to be accurate even in the presence of large head rotations. We propose the first end-to-end system for video-based measurement of heart and breathing rate using a deep convolutional network. The system features a new motion representation based on a skin reflection model and a new attention mechanism using appearance information to guide motion estimation, both of which enable robust measurement under heterogeneous lighting and major motions. Our approach significantly outperforms all current state-of-the-art methods on both RGB and infrared video datasets. Furthermore, it allows spatial-temporal distributions of physiological signals to be visualized via the attention mechanism.
\end{abstract}

%
%
%
%
\renewcommand*{\thefootnote}{\arabic{footnote}}

\begin{figure*}[t]
	\centering\noindent
	\includegraphics[width=0.95\linewidth]{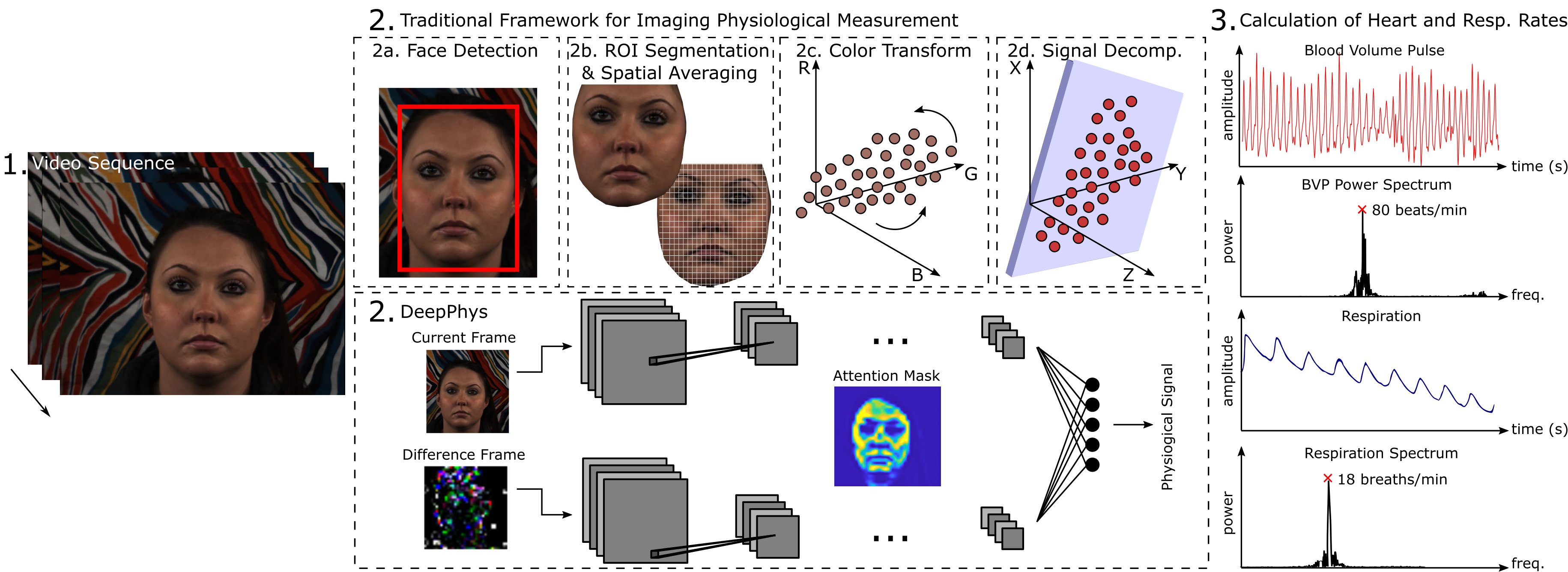}
	\caption{We present DeepPhys a novel approach to video-based physiological measurement using convolutional attention networks that significantly outperforms the state-of-the-art and allows spatial-temporal visualization of physiological information in video. DeepPhys is an end-to-end network that can accurately recover heart rate and breathing rate from RGB or infrared videos.}
	\label{fig:overview}
\end{figure*}

\section{Introduction}
Subtle changes in the human body, ``hidden'' to the unaided eye, can reveal important information about the health and wellbeing of an individual. Computer analysis of videos and images can be used to recover these signals~\cite{takano2007heart,poh2010non,balakrishnan2013detecting} and magnify them~\cite{wu2012eulerian,hurter2017cardiolens}. Non-contact video-based physiological measurement is a fast growing research domain as it offers the potential for unobtrusive, concomitant measurement and visualization of important vital signs using ubiquitous sensors (e.g. low-cost webcams or smartphone cameras)~\cite{mcduff2015survey}. 

Imaging Photoplethysmography (iPPG) is a set of techniques for recovering the volumetric change in blood close to the surface of the skin, the resulting signal is known as the blood volume pulse (BVP). The principle is based on measurement of subtle changes in light reflected from the skin. In a similar fashion, Imaging Ballistocardiography (iBCG) uses motion information extracted from videos to recover the Ballistocardiogram~\cite{balakrishnan2013detecting}. These small motions of the body, due to the mechanical flow of blood, provide complementary cardiac information to the PPG signal. Respiratory signals (breathing) can also be recovered using color~\cite{poh2011advancements} and motion-based~\cite{tarassenko2014non} analyses.  If the physiological parameters (i.e., heart and/or breathing rate) of an individual are known, magnification of the color change and/or motion can be performed~\cite{wu2012eulerian,hurter2017cardiolens}, which provides a very intuitive visualization of the physiological changes.


Based on the techniques above, a number of algorithms have been proposed to enable recovering vital signs from only a webcam video. However, early solutions among them were either not amenable to computation~\cite{takano2007heart,verkruysse2008remote} or lacked rigorous evaluation across a range of motion and ambient lighting conditions~\cite{poh2010non,aarts2013non}. 
Recent approaches that have presented improved results under such video conditions are often complex multi-stage methods that are hard to tune and implement. Almost all methods require face tracking and registration, skin segmentation, color space transformation, signal decomposition and filtering steps.

In computer vision end-to-end deep neural models have out-performed traditional multi-stage methods that require hand-crafted feature manipulation. An end-to-end learning framework for recovering physiological signals would be desirable. However, to date convolution neural networks (CNNs) have only been applied to skin segmentation in iPPG~\cite{chaichulee2017multi} and not for recovering vital signs. This motivated our design of DeepPhys - a novel convolutional attention network (CAN) for video-based physiological measurement, which significantly outperforms the state-of-the-art and allows spatial-temporal visualization of physiological signal distributions (e.g., blood perfusion) from RGB videos. Fig.~\ref{fig:overview} shows how DeepPhys compares to traditional methods.

An end-to-end network for video-based physiological measurement should be a model that reads motion information from video frames, discriminates different motion sources and synthesizes a target motion signal. However, no existing deep learning models are suited to this task. First, deep neural networks used in similar tasks commonly rely on motion representations such as optical flow or frame difference as input, but they are either contradictory to the principle of iPPG or sensitive to different lighting conditions or skin contours. Thus, in DeepPhys, we propose a new motion representation calculating normalized frame difference based on the skin reflection model, which can better capture physiological motions under heterogeneous illumination. Second, the appearance information in video such as the color and texture of the human skin can guide where and how the physiological motions should be estimated, but there is currently no way to jointly learn auxiliary appearance representations along with motion estimators in neural networks. Therefore, we propose a new mechanism that acquires attention from the human appearance to assist the motion learning.



To summarize, we present the first end-to-end method for recovering physiological signals (HR, BR) from videos.  DeepPhys is a novel convolutional attention network that simultaneously learns a spatial mask to detect the appropriate regions of interest and recovers the BVP and respiration signals. Below we describe the theoretical underpinning of our approach, the model, and validation on four datasets each recorded with different devices, of different subjects, and under different lighting configurations. We perform a thorough comparison against the state-of-the-art approaches, the results of which illustrate the benefits of our proposed model. 


\section{Related Work}

\subsection{Remote Physiological Measurement}
Minute variations in light reflected from the skin can be used to extract human physiological signals (e.g., heart rate (HR)~\cite{takano2007heart} and breathing rate (BR)~\cite{poh2011advancements}). 
A digital single reflex camera (DSLR) is sufficient to measure the subtle blood volume pulse signal (BVP)~\cite{takano2007heart,verkruysse2008remote}. 
The simplest method involves spatially averaging the image color values for each frame within a time window, this is highly susceptible to noise from motion, lighting and sensor artifacts. Recent advancements have led to significant improvements in measurement under increasingly challenging conditions. Fig.~\ref{fig:overview} shows a traditional approach to remote physiological measurement that involves skin segmentation, color space transformation and signal decomposition.

\textbf{Color Space Transforms:}
The optical properties of the skin under ambient illumination mean that the green color channel tends to give the strongest PPG signal and this was used in initial work~\cite{verkruysse2008remote,li2014remote}. However, the color channels can be weighted and combined to yield better results. The CHROM~\cite{de2013robust} method uses a linear combination of the chrominance signals by assuming a standardized skin color profile to white-balance the video frames. The Pulse Blood Vector (PBV) method~\cite{de2014improved} leverages the characteristic blood volume changes in different parts of the frequency spectrum to weight the color channels.

\textbf{Signal Decomposition:}
It has been demonstrated that blind-source separation can be used to improve the signal-to-noise ratio of the PPG signal from webcam videos~\cite{poh2010non}. Leveraging information from all three color channels and multiple spatial locations makes the approach more robust to changes in illumination with head motion. More rigorous evaluation demonstrated it was possible to recover heart rate variability (HRV) and breathing rate (BR) estimates~\cite{poh2011advancements}. A majority of the work on video-based physiological measurement has relied on unsupervised learning.  Independent component analysis (ICA)~\cite{poh2010non,poh2011advancements,mcduff2014improvements} and principal component analysis (PCA)~\cite{wang2015exploiting} are two common approaches for combining multiple color or location channels. More advanced signal decomposition methods have resulted in improved heart rate measurement, even in the presence of large head motions and lighting variation~\cite{Wang2016b,wang2016novel,lam2015robust,tulyakov2016self}. 

\textbf{Supervised Learning Approaches:}
Few approaches have made use of supervised learning for video-based physiological measurement.  Formulating the problem is not trivial. Template matching and Support Vector approaches~\cite{osman2015supervised} have obtained modest results.  Linear regression and Nearest Neighbor (NN) techniques have been combined with signal decomposition methods~\cite{monkaresi2014machine} to solve the problem of selecting the appropriate source signal.  However, these are still limited by the performance of the decomposition method (e.g., ICA or PCA).

\subsection{Deep Learning}
\textbf{Motion Analysis} plays a significant role in deep-learning-based video processing. First, deep learning has achieved remarkable successes in explicit motion analysis tasks, such as optical flow estimation~\cite{dosovitskiy2015flownet,Ilg2017} and motion prediction~\cite{Finn2016,Xue2016}. Different from images, videos have both spatial information (appearance) and temporal dynamics (motion). Therefore, solving any video-related problem using machine learning should benefit from implicit motion modeling. In contemporary techniques, both appearance and motion representations can be learned in parallel (two-stream methods~\cite{Simonyan2014}), in cascade (CNN connected with RNN~\cite{Donahue2015,Ballas2016}) or in a mixed way (3D CNN~\cite{Jia2014}). Commonly, for more efficient learning, several motion representations are manually calculated from videos to serve as the input of learning models, including optical flow~\cite{Simonyan2014,Ng} and frame difference~\cite{Xue2016}.  

\textbf{Attention Mechanisms} 
in neural networks are inspired by the human visual system, which can focus on a certain region of an image with high resolution while perceiving the surrounding image in low resolution. To put it simply, they confer more weight on a subset of features. As one of the latest advancements in deep learning, attention mechanisms have been widely used in machine translation~\cite{bahdanau2014neural}, image captioning~\cite{xu2015show} and many other tasks. In learning-based video analytics, attention mechanisms also show great power, either by temporally focusing on different frames of a video~\cite{yao2015video} or by spatially focusing on different parts of a frame~\cite{sharma2015action}. It has been shown that attention can be derived from motions to guide appearance representation learning~\cite{li2018videolstm,tran2017two}. In this work, we do exactly the opposite, acquiring attention from appearance to guide motion representation learning, which has never been done before to our knowledge.

\section{Skin Reflection Model}
Video-based physiological measurement involves capturing both subtle color changes (iPPG) and small motions (iBCG and respiratory movement) of the human body using a camera. For modeling lighting, imagers and physiology, previous works used the Lambert-Beer law (LBL) \cite{lam2015robust,Xu2014a} or Shafer's dichromatic reflection model (DRM) \cite{Wang2016b}. 
We build our learning model on top of the DRM as it provides a better framework for modeling both color changes and motions. Assume the light source has a constant spectral composition but varying intensity. We can define the RGB values of the $k$-th skin pixel in an image sequence by a time-varying function:
\begin{equation} \label{eq:1}
	\pmb{C}_k(t)=I(t) \cdot (\pmb{v}_s(t)+\pmb{v}_d(t))+\pmb{v}_n(t)
\end{equation}
where $\pmb{C}_k(t)$ denotes a vector of the RGB values; $I(t)$ is the luminance intensity level, which changes with the light source as well as the distance between the light source, skin tissue and camera; $I(t)$ is modulated by two components in the DRM: specular reflection $\pmb{v}_s(t)$, mirror-like light reflection from the skin surface, and diffuse reflection $\pmb{v}_d(t)$, the absorption and scattering of light in skin-tissues; $\pmb{v}_n(t)$ denotes the quantization noise of the camera sensor.
$I(t)$, $\pmb{v}_s(t)$ and $\pmb{v}_d(t)$ can all be decomposed into a stationary and a time-dependent part through a linear transformation \cite{Wang2016b}:
\begin{equation} \label{eq:2}
	\pmb{v}_d(t) = \pmb{u}_d \cdot d_0 + \pmb{u}_p \cdot p(t)  
\end{equation}
where $\pmb{u}_d$ denotes the unit color vector of the skin-tissue; $d_0$ denotes the stationary reflection strength; $\pmb{u}_p$ denotes the relative pulsatile strengths caused by hemoglobin and melanin absorption; $p(t)$ denotes the BVP.
\begin{equation} \label{eq:3}
	\pmb{v}_s(t) = \pmb{u}_s \cdot (s_0+\Phi(m(t),p(t))) 
\end{equation}
where $\pmb{u}_s$ denotes the unit color vector of the light source spectrum; $s_0$ and $\Phi(m(t),p(t))$ denote the stationary and varying parts of specular reflections; $m(t)$ denotes all the non-physiological variations such as flickering of the light source, head rotation and facial expressions.  
\begin{equation} \label{eq:4}
	I(t) = I_0 \cdot (1+\Psi(m(t),p(t))) 
\end{equation}
where $I_0$ is the stationary part of the luminance intensity, and $I_0\cdot\Psi(m(t),p(t))$ is the intensity variation observed by the camera.
The interaction between physiological and non-physiological motions, $\Phi(\cdot)$ and $\Psi(\cdot)$, are usually complex non-linear functions.
The stationary components from the specular and diffuse reflections can be combined into a single component representing the stationary skin reflection:
\begin{equation} \label{eq:5}
	\pmb{u}_c \cdot c_0 = \pmb{u}_s \cdot s_0 + \pmb{u}_d \cdot d_0
\end{equation}
where $\pmb{u}_c$ denotes the unit color vector of the skin reflection and $c_0$ denotes the reflection strength. Substituting (\ref{eq:2}), (\ref{eq:3}), (\ref{eq:4}) and (\ref{eq:5}) into (\ref{eq:1}), produces:
\begin{equation} \label{eq:6}
	\pmb{C}_k(t)=I_0\cdot (1+\Psi(m(t),p(t))) \cdot \\
	(\pmb{u}_c \cdot c_0+\pmb{u}_s \cdot \Phi(m(t),p(t))+\pmb{u}_p \cdot p(t))+\pmb{v}_n(t)
\end{equation}
As the time-varying components are much smaller (i.e., orders of magnitude) than the stationary components in (\ref{eq:6}), we can neglect any product between varying terms and approximate $\pmb{C}_k(t)$ as:
\begin{equation} \label{eq:7}
	\pmb{C}_k(t)\approx \pmb{u}_c \cdot I_0 \cdot c_0+\pmb{u}_c \cdot I_0 \cdot c_0 \cdot \Psi(m(t),p(t)) + \\
	\pmb{u}_s \cdot I_0 \cdot \Phi(m(t),p(t))+\pmb{u}_p \cdot I_0 \cdot p(t)+\pmb{v}_n(t)
\end{equation}

For any of the video-based physiological measurement methods, the task is to extract $p(t)$ from $\pmb{C}_k(t)$. To date, all iPPG works have ignored $p(t)$ inside $\Phi(\cdot)$ and $\Psi(\cdot)$, and assumed a linear relationship between $\pmb{C}_k(t)$ and $p(t)$. This assumption generally holds when $m(t)$ is small (i.e., the skin ROI is stationary under constant lighting conditions). 
However, $m(t)$ is not small in most realistic situations. Thus, a linear assumption will harm measurement performance. This motivates our use of a machine learning model to capture the more general and complex relationship between $\pmb{C}_k(t)$ and $p(t)$ in (\ref{eq:7}).

\section{Approach}
\subsection{Motion Representation}
We developed a new type of normalized frame difference as our input motion representation.
Optical flow, though commonly used, does not fit our task, because it is based on the brightness constancy constraint, which requires the light absorption of objects to be constant. This obviously contradicts the existence of a varying physiological signal $p(t)$ in (\ref{eq:2}). 
The first step of computing our motion representation is
spatial averaging of pixels, which has been widely used for reducing the camera quantization error $\pmb{v}_n(t)$ in (\ref{eq:7}). We implemented this by downsampling every frame to $L$ pixels by $L$ pixels using bicubic interpolation. Selecting $L$ is a trade-off between suppressing camera noise and retaining spatial resolution (\cite{wang2015exploiting} found that $L = 36$ was a good choice for face videos.) The downsampled pixel values will still obey the DRM model only without the camera quantization error:
\begin{equation} \label{eq:8}
	\pmb{C}_l(t)\approx \pmb{u}_c \cdot I_0 \cdot c_0+\pmb{u}_c \cdot I_0 \cdot c_0 \cdot \Psi(m(t),p(t)) + \\
	\pmb{u}_s \cdot I_0 \cdot \Phi(m(t),p(t))+\pmb{u}_p \cdot I_0 \cdot p(t)
\end{equation}
where $l=1,\cdots,L^2$ is the new pixel index in every frame.

\begin{figure*}[t]
	\centering\noindent
	\includegraphics[width=\linewidth]{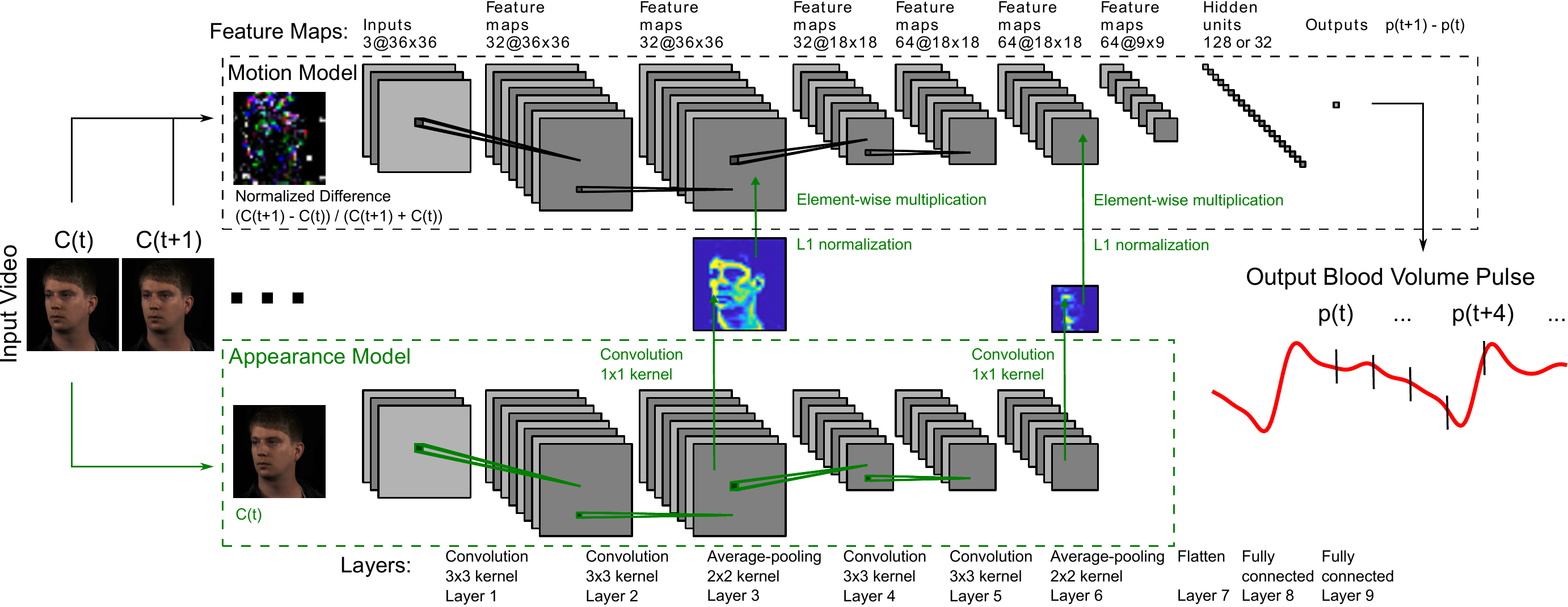}
	\caption{The architecture of our end-to-end convolutional attention network. The current video frame at time $\textit{t}$ and the normalized difference between frames at $\textit{t+1}$ and $\textit{t}$ are given as inputs to the appearance and motion models respectively. The network learns spatial masks, that are shared between the models, and features important for recovering the BVP and respiration signals.}
	\label{fig:network}
\end{figure*}

Then we need to reduce the dependency of $\pmb{C}_l(t)$ on the stationary skin reflection color $\pmb{u}_c \cdot I_0 \cdot c_0$, resulting from the light source and subject's skin tone. In unsupervised learning approaches, the frames processed usually come from a short time window, in which the term $\pmb{u}_c \cdot I_0 \cdot c_0$ is relatively constant.
However, in a supervised learning data cohort, the term will vary between subjects and lighting conditions, which will explain the majority of the variance in $\pmb{C}_l(t)$. This will not only make it harder to learn to discriminate the real variance of interest $p(t)$, but also depend the learned model on specific skin tones and lamp spectra in the training data. In (\ref{eq:8}) $\pmb{u}_c \cdot I_0 \cdot c_0$ appears twice. It is impossible to eliminate the second term as it interacts with $\Psi(\cdot)$. However, the first time-invariant term, which is usually dominant, can be removed by taking the first order derivative of both sides of (\ref{eq:8}) with respect to time:
\begin{equation} \label{eq:9}
	\pmb{C}'_l(t)\approx \pmb{u}_c \cdot I_0 \cdot c_0 \cdot (\frac{\partial\Psi}{\partial m}m'(t) + \frac{\partial\Psi}{\partial p}p'(t)) + \\
	\pmb{u}_s \cdot I_0 \cdot (\frac{\partial\Phi}{\partial m}m'(t) + \frac{\partial\Phi}{\partial p}p'(t))+\pmb{u}_p \cdot I_0 \cdot p'(t)
\end{equation}


One problem with this frame difference representation is that the stationary luminance intensity level $I_0$ is spatially heterogeneous due to different distances to the light source and uneven skin contours. The spatial distribution of $I_0$ has nothing to do with physiology, but is different in every video recording setup. Thus, $\pmb{C}'_l(t)$ was normalized by dividing it by the temporal mean of $\pmb{C}_l(t)$ to remove $I_0$:
\begin{multline} \label{eq:10}
	\frac{\pmb{C}'_l(t)}{\overline{\pmb{C}_l(t)}}\approx \pmb{1} \cdot (\frac{\partial\Psi}{\partial m}m'(t) + \frac{\partial\Psi}{\partial p}p'(t)) +diag^{-1}(\pmb{u}_c)\pmb{u}_p \cdot \frac{1}{c_0} \cdot p'(t)+ \\
	diag^{-1}(\pmb{u}_c)\pmb{u}_s \cdot \frac{1}{c_0} \cdot (\frac{\partial\Phi}{\partial m}m'(t) + \frac{\partial\Phi}{\partial p}p'(t))
\end{multline}
where $\pmb{1}=[1~1~1]^T$. In (\ref{eq:10}), $\overline{\pmb{C}_l(t)}$ needs to be computed pixel-by-pixel over a short time window to minimize occlusion problems and prevent the propagation of errors. We found it was feasible to compute it over two consecutive frames so that (\ref{eq:10}) can be expressed discretely as:
\begin{equation} \label{eq:11}
	\pmb{D}_l(t) = \frac{\pmb{C}'_l(t)}{\overline{\pmb{C}_l(t)}}\sim\frac{\pmb{C}_l(t+\Delta t)-\pmb{C}_l(t)}{\pmb{C}_l(t+\Delta t)+\pmb{C}_l(t)}
\end{equation}
which is the normalized frame difference we used as motion representation ($\Delta t$ is the sampling interval).
In the computed $\pmb{D}_l(t)$, outliers are usually due to large $m'(t)$ or occlusion. To diminish these outliers, we clipped $\pmb{D}_l(t)$ by three standard deviations over each video and all color channels.
To summarize, the clipped $\pmb{D}_l(t)$ will be the input of our learning model, and the first order derivative of a gold-standard physiological signal $p'(t)=p(t+\Delta t)-p(t)$ will be the training label. To align $\pmb{D}_l(t)$ and $p'(t)$, the physiological signals were interpolated to the video sampling rate beforehand using piecewise cubic Hermite interpolation. For higher convergence speed of stochastic gradient descent, $\pmb{D}_l(t)$ and $p'(t)$ were also scaled to unit standard deviation over each video.


\subsection{Convolutional Neural Network}
The foundation of our learning model is a VGG-style CNN for estimating the physiological signal derivative from the motion representation, as shown in Fig. \ref{fig:network} (motion model).  The last layer has linear activation units and a mean square error (MSE) loss, so the outputs will form a continuous signal $\widetilde{p'(t)}$. Since most physiological signals are frequency-bounded, the output was band-pass filtered to remove noise outside the frequency range of interest. 
Finally, the power spectrum was computed from the filtered signal with the location of the highest peak taken as the estimated HR or BR.

Different from classical CNN models deigned for object recognition, we used average pooling layers instead of max pooling. The reasoning was that for physiological measurement, combining an important feature with a less important feature can often produce a higher signal-to-noise ratio than using the more important feature alone.
We also compared multiple activation functions and found the symmetry seemed to help with performance. Thus, we used hyperbolic tangent (tanh) instead of rectified linear units (ReLU) as hidden layer activation functions. 
Besides, our attention mechanism uses gating schemes similar to those in long short-term memory (LSTM) networks, which help prevent the main problem with tanh, vanishing gradients.

The loss function of our model is the MSE between the estimated and gold-standard physiological signal derivative, but our final goal is to compute the dominant frequency of the estimated signal (i.e., HR or BR). Though the temporal error and frequency error generally have high correlation, a small temporal error does not guarantee a small frequency error. It is also hard to directly use the frequency error as the loss function of a CNN, because the calculation of the dominant frequency involves a non-differentiable operation $argmax$. 
Thus we adopted ensemble learning over training checkpoints. 
Specifically, we trained CNN models for an extra 16 epochs after convergence. These models were applied to the training data with frequency errors computed, and the model with the smallest error was chosen. We found that this strategy consistently achieved smaller frequency errors than simply using the last checkpoint model.

\subsection{Attention Mechanism}
Naively our motion representation in (\ref{eq:10}) and (\ref{eq:11}) assumes every pixel $l$ is part of a body, and more specifically skin. Using normalized frame difference helps reduce the impact of background pixels; however, any movement will add noise. To reduce this effect, previous methods usually reply on preprocessing such as skin segmentation to select a region of interest (ROI). In our end-to-end model, a new attention mechanism can be added to achieve similar functionality. Also, the distribution of physiological signals is not uniform on the human body, so learning soft-attention masks and assigning higher weights to skin areas with stronger signals should improve the measurement accuracy.

Whether a pixel patch belongs to the skin and exhibits strong physiological signals can be partly inferred from its visual appearance. However, the derivation and normalization operations in (\ref{eq:10}) removed appearance information. To provide a basis for learning attention, we created a separate appearance model (see Fig.~\ref{fig:network}). This model has the same architecture as the motion model without the last three layers, and has the raw frames (centered to zero mean and scaled to unit standard deviation) as input. Soft-attention masks were estimated using a $1\times 1$ convolution filter right before every pooling layer so that masks were synthesized from different levels of appearance features. Let $\mathbbm{x}_m^j\in \mathbb{R}^{C_j\times H_j\times W_j}$ and $\mathbbm{x}_a^j\in \mathbb{R}^{C_j\times H_j\times W_j}$ be the feature maps of convolution layer $j$ right before pooling in the motion model and the appearance model respectively, with $C_j$, $H_j$ and $W_j$ being the number of channels, height and width. The attention mask $\mathbbm{q}^j\in \mathbb{R}^{1\times H_j\times W_j}$ can be computed as:
\begin{equation}
	\mathbbm{q}^j = \frac{H_j W_j\cdot\sigma({\pmb{w}^j}^T\mathbbm{x}_a^j+b^j)}{2\lVert\sigma({\pmb{w}^j}^T\mathbbm{x}_a^j+b^j)\rVert_1}
\end{equation}
where $\pmb{w}^j\in \mathbb{R}^{C_j}$ is the $1\times 1$ convolution kernel, $b^j$ is the bias, and $\sigma(\cdot)$ is a sigmoid function. Different from softmax functions commonly used for generating soft-attention probability maps, we used a sigmoid activation followed by $l_1$ normalization, which is even softer than softmax and produces less extreme masks. Finally, the mask was multiplied with the motion model feature map to output:
\begin{equation}
	\mathbbm{z}_m^j = (\pmb{1}\cdot \mathbbm{q}^j) \odot \mathbbm{x}_m^j 
\end{equation}
where $\mathbbm{z}_m^j\in \mathbb{R}^{C_j\times H_j\times W_j}$ is the masked feature map passed on to the next layer, $\pmb{1}\in \mathbb{R}^{C_j}$ is a vector with all ones, and $\odot$ is element-wise multiplication. The motion model and the appearance model were learned jointly to find the best motion estimator and the best ROI detector simultaneously.

\section{Datasets}
We tested our method on four datasets, each featuring participants of both genders, different ages, a wide range of skin tones (Asians, Africans and Caucasians) and some had thick facial hair and/or glasses. 

\textbf{RGB Video I \cite{Estepp}.}
Videos were recorded with a Scout scA640-120gc GigE-standard, color camera, capturing 8-bit, 658x492 pixel images, 120 fps. The camera was equipped with 16 mm fixed focal length lenses. Twenty-five participants (17 males) were recruited to participate for the study.
Gold-standard physiological signals were measured using a research-grade biopotential acquisition unit.

Each participant completed six (each against two background screens) 5-minute tasks.  The tasks were designed to capture different levels of head motion. 
\newline
\textbf{Task 1:} Participants placed their chin on a chin rest (normal to the camera) in order to limit head motion.
\newline
\textbf{Task 2:} Participants repeated Task 1 without the aid of the chin rest, allowing for small natural motions.
\newline
\textbf{Task 3:} Participants performed a 120-degree sweep centered about the camera at a speed of 10 degrees/sec.
\newline
\textbf{Task 4:} As Task 3 but with a speed of 20 degrees/sec.
\newline
\textbf{Task 5:} As Task 3 but with a speed of 30 degrees/sec.
\newline
\textbf{Task 6:} Participants were asked to reorient their head position once per second to a randomly chosen imager in the array. Thus simulating random head motion.

\begin{figure}[t]
	\centering\noindent
	\includegraphics[width=0.95\linewidth]{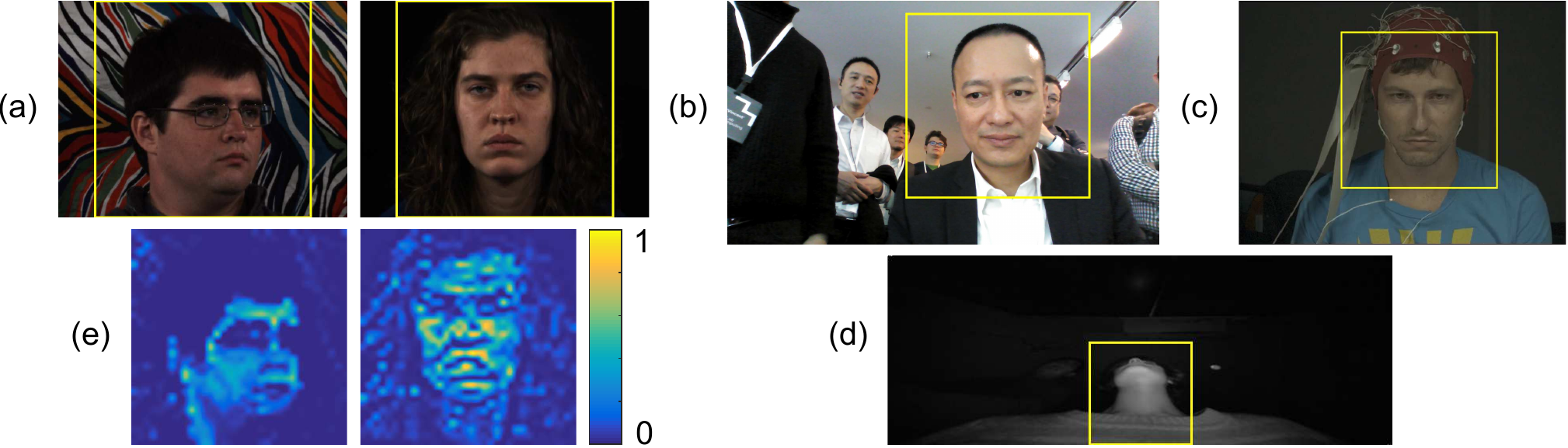}
	\caption{Example frames from the four datasets: (a) RGB Video I, (b) RGB Video II, (c) MAHNOB-HCI, (d) Infrared Video. The yellow bounding boxes indicate the areas cropped as the input of our models. (e) shows exemplary attention weights of the left frame in (a) for HR measurement, and of the right frame in (a) for BR measurement.}
	\label{fig:example_frames}   
\end{figure}

\textbf{RGB Video II \cite{chen2017eliminating}.}
Videos were recorded with an Intel RealSense Camera VF0800 and 18 participants (16 males, 2 females, 23-50 years) were recruited (data from 3 participants were eliminated due to high collection error).
Participant was seated still at a desk under ambient light for 30 s. All videos were recorded in color (24-bit RGB with 3 channels × 8 bits/channel) at a floating frame rate around 24 fps with pixel resolution of 1920x1080. Gold-standard physiological signals were measured with a FlexComp Infiniti that recorded Blood Volume Pulse (BVP) from a finger probe at a constant sampling frequency of 256 Hz. Respiration was not recorded in this study.

\textbf{MAHNOB-HCI \cite{soleymani2012multimodal}.}
A multimodal dataset with videos of subjects participating in two experiments: (i) emotion elicitation and (ii) implicit tagging. It contains 27 subjects (12 males and 15 females) in total, and all videos are in 61 fps with a 780x580 resolution.  Following \cite{li2014remote}, we used a 30-second clip (frames from 306 through 2135) of 527 sequences. To compute the ground truth heart rate we used the second channel (EXG2) of the corresponding ECG waveforms.

\textbf{Infrared Video \cite{WeixuanChen2016}.}
To show that our approach generalizes to other datasets collected using different imagers we performed similar analysis on a dataset of IR videos. 
Twelve participants (8 males, 4 females) between the ages of 23-34 years were recruited for this study.
Participants were seated at a desk and a Leap Motion controller was placed on the edge of the desk, parallel with it, and facing upward. 640x240 pixel near-IR frames were recorded at a floating frame rate around 62 fps.
Gold-standard physiological signals were measured with a FlexComp Infiniti that simultaneously recorded BVP from a finger probe and respiration from a chest belt at a constant sampling frequency of 256 Hz. 
Each experiment consisted of two 1 min recordings: 1) in a well-lit room (184 lux at the camera), 2) in a completely dark room (1 lux at the camera).  

\begin{table*}[!ht]
	\centering\noindent
	\caption{Performance of heart rate and breathing rate measurement for RGB Video I. Participant dependent (p.-dep.) and participant independent results are shown, as are task independent results for the six tasks with varying levels of head rotation}
	\label{tab:respiration}
	\footnotesize
	\setlength\tabcolsep{3pt}
	\begin{tabular}{p{2.8cm}|ccccccc|ccccccc}
		
		\textbf{HEART RATE} & \multicolumn{7}{c|}{Mean Absolute Error / BPM} & \multicolumn{7}{c}{Signal-To-Noise Ratio / dB} \\
		Methods & 1 & 2 & 3 & 4 & 5 & 6 & Avg. & 1 & 2 & 3 & 4 & 5 & 6 & Avg. \\
		\hline\hline
		Estepp et al.~\cite{Estepp} & 3.48 &    3.95 & 3.80 & 6.55 &  11.8 & 13.4 & 7.16 & 6.06 & 4.82 & 3.77 & -0.10 & -4.72 & -9.63 & 0.03 \\
		McDuff et al.~\cite{mcduff2014improvements} & 1.17 & 1.70 & 1.70 & 4.00 & 5.22 & 11.8 & 4.29 & 10.9 & 9.55 & 6.69 & 3.08 & 0.08 & -6.93 & 3.90 \\
		Balakrishnan et al.~\cite{balakrishnan2013detecting} & 4.99 & 5.16 & 12.7 & 17.4 & 18.7 & 14.2 & 12.2 & -1.08 & -0.34 & -8.83 & -12.6 & -14.2 & -12.1 & -8.19 \\
		De Haan et al.~\cite{de2013robust} & 4.53 & 4.59 & 4.35 & 4.84 & 6.89 & 10.3 & 5.92 & 1.72 & 1.38 & 3.97 & 3.63 & 2.02 & -2.47 & 1.71 \\
		Wang et al.~\cite{Wang2016b} & 1.50 & 1.53 & 1.50 & 1.84 & 2.05 & 6.11 & 2.42 & 6.84 & 6.21 & 4.80 & 2.97 & 0.77 & -4.33 & 2.88 \\
		Tulyakov et al.~\cite{tulyakov2016self} & 1.76 & 2.14 & 14.9 & 19.0 & 15.7 & 22.0 & 12.6 & 4.32 & 2.29 & -11.8 & -14.3 & -12.3 & -15.3 & -7.85 \\
		\hline\hline
		\textbf{OURS: Part. Dep.} \\
		Motion-only CNN & 1.17 & 1.29 & 1.29 & 1.66 & 2.04 & 2.95 & 1.73 & 10.2 & 9.28 & 7.02 & 4.18 & 1.95 & -1.00 & 5.28 \\
		Stacked CNN & 1.18 & 1.30 & 1.26 & 1.51 & 1.82 & 2.62 & 1.61 & 10.7 & 9.60 & 8.28 & 5.59 & 2.84 & 0.29 & 6.22 \\
		CAN & 1.17 & 1.26 & 1.16 & 1.61 & 2.04 & 1.78 & 1.50 & 10.9 & 9.66 & 8.20 & 5.69 & 3.57 & 1.33 & 6.55 \\
		\hline\hline
		\textbf{OURS: Part. Ind.} \\
		Motion-only CNN & 1.17 & 1.41 & 1.33 & 1.91 & 3.57 & 6.41 & 2.63 & 9.60 & 8.62 & 6.25 & 2.73 & -0.15 & -4.02 & 3.84 \\
		Stacked CNN & 1.12 & 1.66 & 1.30 & 1.65 & 2.33 & 6.25 & 2.38 & 9.26 & 8.18 & 7.27 & 3.81 & 1.14 & -2.84 & 4.47 \\
		CAN & 1.16 & 1.45 & 1.23 & 1.72 & 2.42 & 5.59 & 2.26 & 9.52 & 8.82 & 7.36 & 3.90 & 1.10 & -2.72 & 4.66 \\
		\hline
		CAN (task 1) & 1.16 & 3.01 & 3.43 & 4.85 & 7.92 & 13.9 & 5.70 & 9.52 & 5.74 & 2.01 & -2.13 & -5.77 & -9.70 & -0.06 \\
		CAN (task 2) & 1.12 & 1.45 & 1.67 & 3.35 & 7.51 & 12.9 & 4.66 & 10.1 & 8.82 & 4.25 & -0.90 & -5.33 & -8.96 & 1.33 \\
		CAN (task 3) & 1.18 & 1.41 & 1.23 & 1.78 & 2.73 & 9.23 & 2.93 & 9.75 & 8.79 & 7.36 & 4.33 & 0.55 & -5.88 & 4.15 \\
		CAN (task 4) & 1.13 & 1.57 & 1.24 & 1.72 & 2.79 & 8.98 & 2.91 & 9.95 & 8.72 & 7.06 & 3.90 & 0.37 & -5.68 & 4.05 \\
		CAN (task 5) & 1.16 & 1.38 & 1.30 & 1.54 & 2.42 & 7.05 & 2.48 & 9.86 & 8.98 & 7.68 & 4.40 & 1.10 & -4.15 & 4.65 \\
		CAN (task 6) & 1.14 & 1.32 & 1.22 & 1.47 & 2.17 & 5.59 & 2.15 & 10.4 & 9.62 & 8.21 & 5.15 & 1.83 & -2.72 & 5.41 \\
		CAN (all tasks) & 1.13 & 1.34 & 2.45 & 1.64 & 1.83 & 6.32 & 2.45 & 9.88 & 7.21 & 1.46 & 8.48 & 4.11 & -3.23 & 4.65 \\
		
	\end{tabular}
	\qquad 
	\begin{tabular}{p{2.8cm}|ccccccc|ccccccc}
		
		\textbf{BREATH. RATE} & \multicolumn{7}{c|}{Mean Absolute Error / BPM} & \multicolumn{7}{c}{Signal-To-Noise Ratio / dB} \\
		Methods & 1 & 2 & 3 & 4 & 5 & 6 & Avg. & 1 & 2 & 3 & 4 & 5 & 6 & Avg. \\
		\hline\hline
		Tarassenko et al.~\cite{tarassenko2014non} & 2.51 & 2.53 & 3.19 & 4.85 & 4.22 & 4.78 & 3.68 & -1.29 & -1.82 & -6.32 & -8.55 & -8.79 & -10.6 & -6.22 \\
		\hline\hline
		\textbf{OURS: Part. Dep.} \\
		Motion-only CNN & 2.03 & 2.47 & 3.21 & 3.04 & 3.11 & 4.27 & 3.02 & -0.33 & -1.91 & -5.28 & -4.83 & -5.33 & -9.64 & -4.55 \\
		Stacked CNN & 1.74 & 2.27 & 2.98 & 2.79 & 3.03 & 5.33 & 3.02 & 1.84 & -0.93 & -6.31 & -5.18 & -5.70 & -11.2 & -4.58 \\
		CAN & 1.70 & 2.19 & 3.24 & 3.05 & 3.06 & 3.96 & 2.86 & 2.73 & -0.02 & -4.39 & -4.47 & -4.36 & -7.97 & -3.08 \\
		\hline\hline
		\textbf{OURS: Part. Ind.} \\
		Motion-only CNN & 1.70 & 2.31 & 4.09 & 4.85 & 4.60 & 4.06 & 3.60 & 0.75 & -0.17 & -6.03 & -9.19 & -9.05 & -9.06 & -5.46 \\
		Stacked CNN & 2.00 & 2.10 & 5.67 & 5.55 & 6.34 & 5.76 & 4.57 & 0.19 & -0.25 & -12.0 & -11.5 & -12.7 & -13.0 & -8.20 \\
		CAN & 1.28 & 1.64 & 4.15 & 4.37 & 3.77 & 4.37 & 3.26 & 4.45 & 2.96 & -5.05 & -6.72 & -6.70 & -8.93 & -3.33 \\
		\hline
		CAN (task 1) & 1.28 & 1.72 & 6.34 & 7.28 & 6.28 & 4.01 & 4.48 & 4.45 & 2.37 & -10.0 & -13.5 & -13.9 & -9.02 & -6.60 \\
		CAN (task 2) & 1.21 & 1.64 & 5.73 & 5.65 & 4.92 & 3.82 & 3.83 & 4.39 & 2.96 & -7.97 & -12.3 & -11.7 & -8.54 & -5.52 \\
		CAN (task 3) & 1.62 & 1.71 & 4.15 & 4.57 & 4.16 & 3.56 & 3.30 & 3.38 & 2.74 & -5.05 & -7.05 & -6.71 & -6.23 & -3.15 \\
		CAN (task 4) & 1.74 & 1.85 & 3.80 & 4.37 & 4.60 & 3.42 & 3.30 & 2.69 & 2.56 & -5.10 & -6.72 & -7.86 & -6.63 & -3.51 \\
		CAN (task 5) & 1.65 & 1.74 & 4.37 & 4.45 & 3.77 & 3.37 & 3.22 & 2.90 & 3.18 & -6.20 & -7.02 & -6.70 & -5.69 & -3.25 \\
		CAN (task 6) & 2.06 & 1.76 & 5.89 & 5.92 & 5.21 & 4.37 & 4.20 & 1.11 & 1.54 & -9.72 & -11.9 & -10.1 & -8.93 & -6.33 \\
		CAN (all tasks) & 1.54 & 4.69 & 3.71 & 2.14 & 5.27 & 3.33 & 3.45 & 3.09 & -6.45 & -7.09 & 0.21 & -10.7 & -7.36 & -4.71 \\
	\end{tabular}
\end{table*}

\section{Results and Discussion}
For our experiments, we used Adadelta~\cite{zeiler2012adadelta} to optimize the models across a computing cluster. All the optimizer parameters were copied from~\cite{zeiler2012adadelta}, and a batch size of 128 examples was used. To overcome overfitting, three dropout layers~\cite{Srivastava2014} were inserted between layers 3 and 4, layers 6 and 7, and layers 8 and 9 in Fig.~\ref{fig:network} with dropout rates $d_1$, $d_2$ and $d_3$ respectively. Along with $n_8$ the number of hidden units in layer 8 and $N_e$ the number of training epochs, the five parameters were chosen differently to adapt to different model complexities and different generalization challenges (values can be found in Supplemental Materials). 

Apart from the proposed CAN model, we also implemented a standalone CNN motion model (top portion of Fig.~\ref{fig:network}) to verify the effectiveness of the attention mechanism. The input of the model was either the normalized frame difference (Motion-only CNN) or the normalized frame difference stacked with the raw frame (Stacked CNN). A 6th-order Butterworth filter was applied to the model outputs (cut-off frequencies of 0.7 and 2.5 Hz for HR, and 0.08 and 0.5 Hz for BR). The filtered signals were divided into 30-second windows with 1-second stride and four standard metrics were computed over all windows of all the test videos in a dataset: mean absolute error (MAE), root mean square error (RMSE), Pearson's correlation coefficient ($r$) between the estimated HR/BR and the ground truth HR/BR, and signal-to-noise ratio (SNR) of the estimated physiological signals \cite{de2013robust} averaged among all windows. The SNR is calculated in the frequency domain as the ratio between the energy around the first two harmonics (0.2 Hz frequency bins around the gold-standard HR, and 0.05 Hz frequency bins around the gold-standard BR) and remaining frequencies within a range of [0.5 4] Hz for HR, and a range of [0.05 1] Hz for BR. Due to limited space, the metrics RMSE and $r$ are shown in Supplemental Materials.

\subsection{RGB Video I}

Every video frame was center-cropped to 492x492 pixels to remove the lateral blank areas before being fed into our processing pipeline. We compare our proposed approach to six other methods~\cite{Estepp,mcduff2014improvements,balakrishnan2013detecting,de2013robust,wang2015exploiting,tulyakov2016self} for recovering the blood volume pulse. For recovering the respiration, we compare to the approach proposed by Tarassenko et al.~\cite{tarassenko2014non}. The details about the implementation of these methods are included in Supplemental Materials.


\textbf{Participant-dependent Performance}
Each five-minute video was divided into five folds of one minute duration. We trained and tested via cross-validation on the concatenated five folds within each task, in which case every participant appeared in both the training set and the test set. The evaluation metrics MAE and SNR are averaged over five folds and shown in Tables~\ref{tab:respiration}.
Table~\ref{tab:respiration} shows that our motion-only CNN, stacked CNN and CAN all outperform the prior methods for HR measurement over task two to task six, both in terms of MAE and SNR. The benefit is particularly strong for the tasks involving high velocity head motions. On task one, our MAE and SNR are very close to the best results achieved by previous methods using hand-crafted features, probably because task one simulates an ideal situation and there is nearly no space for improvement. Within each of our three approaches, CAN shows superior performance on average and obvious advantages for task six, which can be explained by the effectiveness of the attention mechanism in dealing with the frequently changing ROI. The breathing rate results (Table~\ref{tab:respiration}) follow a similar pattern.


\textbf{Participant-independent Performance.}
All the 25 participants were randomly divided into five folds of five participants each. The learning models were trained and tested via five-fold cross-validation within each task to evaluate how our models can be generalized to new participants. The evaluation metrics MAE and SNR are also averaged over five folds and shown in Table~\ref{tab:respiration}.
Compared with the participant-dependent results, the participant-independent results have lower performance to varying degrees. However, for heart rate measurement the Stacked CNN and CAN still outperform all the previous methods. For breathing rate measurement, though motion-only CNN and the stacked CNN have accuracies similar or inferior to Tarassenko et al.~\cite{tarassenko2014non}, the CAN still shows improvement in five tasks and overall. 

\textbf{Task-independent Performance.}
Both the participant-dependent and par-ticipant-independent results are from training and testing models within tasks. Next, we present task-independent performance where the CAN model was trained on a specific task and then tested on other tasks. The training set and test set were again participant-independent. In the HR results shown in Table~\ref{tab:respiration}, there is a clear pattern that a model trained on tasks with less motion performs badly on tasks with greater motion. Models trained on tasks with greater motion generalize very well across all tasks. The CAN model trained on task six even has lower MAE and higher SNR than the model trained and tested within each single task. This also explains why the model trained on all the tasks achieves moderate performance, slightly better than the task five model but much worse than the task six model. On the other hand, for breathing rate measurement, a model trained on one task usually performs best on the same task, and does not generalize well to different tasks. As a result, the distributions of the average MAE and SNR in Table~\ref{tab:respiration} exhibit a symmetric pattern from the task one model to the task six model. 

\begin{table}[!t]
	\caption{RGB Video II, MAHNOB-HCI and Infrared Video dataset results. (MAE = Mean Absolute Error, SNR = Signal-To-Noise Ratio)}
	\label{tab:ir_results}
	\footnotesize
	\setlength\tabcolsep{3pt}
	\begin{minipage}{0.5\textwidth}
		\centering
		\begin{tabular}{p{3cm}|cc|cc}
			\textbf{DATASET} & \multicolumn{2}{c|}{RGB VIDEO II} & \multicolumn{2}{c}{MANHOB-HCI} \\
			& \multicolumn{2}{c|}{Heart Rate} & \multicolumn{2}{c}{Heart Rate} \\
			\multirow{2}{*}{Methods} & MAE & SNR & MAE & SNR  \\
			& /BPM & /dB & /BPM & /dB  \\
			\hline\hline
			Estepp et al.~\cite{Estepp} & 14.7 & -13.2 & - & - \\
			McDuff et al.~\cite{mcduff2014improvements} & 0.25 & -4.48 & 10.5 & -10.4 \\
			Balakrishnan et al.~\cite{balakrishnan2013detecting} & 11.3 & -9.17 & 17.7 & -12.9 \\
			De Haan et al.~\cite{de2013robust} & 0.30 & -2.30 & 5.09 & -9.12 \\
			Wang et al.~\cite{Wang2016b} & 0.26 & 1.50 & - & - \\
			Tulyakov et al.~\cite{tulyakov2016self} & 2.27 & -0.20 & 4.96 & -8.93 \\
			\hline
			\textbf{OURS: Transfer Learning} &  &  \\
			CAN & 0.14 & 0.03 & 4.57 & -8.98 \\
		\end{tabular}
		
	\end{minipage}
	\begin{minipage}{0.5\textwidth}
		\centering
		\begin{tabular}{p{2.7cm}|cc|cc}
			\textbf{DATASET} & \multicolumn{4}{c}{IR Video}  \\
			&  \multicolumn{2}{c|}{Heart Rate} & \multicolumn{2}{c}{Breath. Rate} \\
			\multirow{2}{*}{Methods} & MAE & SNR & MAE & SNR  \\
			& /BPM & /dB & /BPM & /dB  \\            
			\hline\hline
			Chen et al.~\cite{WeixuanChen2016} & 0.65 & 3.15 & 0.27 & 5.71 \\
			\hline
			\textbf{OURS: Part. Ind.} &  &  &  &  \\
			Motion-only CNN & 1.44 & 9.55 & 0.49 & 8.95 \\
			Stacked CNN & 0.87 & 10.9 & 0.14 & 10.4 \\
			CAN & 0.55 & 13.2 & 0.14 & 10.8 \\
			\hline
		\end{tabular}            
	\end{minipage}
\end{table}

\subsection{RGB Video II and MAHNOB-HCI}
As shown in Fig.~\ref{fig:example_frames}, the video frames in the two datasets have complicated backgrounds, and the facial ROI only occupies a small area. To ensure a sufficient number of physiology-related pixels after downsampling, a
face detector based on OpenCV’s Haar-like cascades~\cite{viola2001rapid} was applied to the first frame of each video, and a square region with 160\% width and height of the detected bounding box was cropped as the input to our approach. 

\textbf{Transfer learning.}
To test whether our model can be generalized to videos with a different resolution, background, lighting condition and sampling frequency, we tried transfer learning without any fine-tuning. We trained the models on Task 2 of RGB Video I (the most similar task to RGB Video II and MAHNOB-HCI) and applied them directly to these datasets. Since RGB Video II only has blood volume pulse ground truth, we compare our approach with only the HR measurement methods. For MAHNOB-HCI, as it is public, we evaluated our approach against only those methods reported in previous studies on the dataset. The results are presented in Table~\ref{tab:ir_results}. Without any prior knowledge about the two datasets, our CAN model still attains the lowest MAE compared with any previous method, and its SNR is only second to Wang et al. \cite{wang2015exploiting} on RGB Video II and Tulyakov et al. \cite{tulyakov2016self} on MAHNOB-HCI.

\subsection{Infrared Video}
For these videos we cropped a fixed 130x130 pixels bounding box against the lower boundary of the frame as our model input (see Fig.~\ref{fig:example_frames}d). 
Since the frames in the dataset are monochromatic, all the previous methods we implemented for the RGB datasets are not applicable. We compare our approach with a PCA-based algorithm \cite{WeixuanChen2016}, which has achieved the highest accuracy on the dataset, for both HR and BR measurement.

\textbf{Participant-independent Performance.}
In the dataset, each video is one minute in length, which is too short to be split into multiple folds for participant-dependent evaluation. Thus we only ran experiments in a participant-independent way: The 13 participants were randomly divided into five folds, and the learning models were trained and tested via five-fold cross-validation. The results are averaged over five folds and shown in Table~\ref{tab:ir_results}. For both heart rate and breathing rate measurement, the CAN model not only beats the previous best results but also beats the other learning-based methods without an attention mechanism.

\subsection{Visualization of Attention Weights}
An advantage of utilizing the proposed attention mechanism is that the spatial-temporal distributions of physiological signals can be revealed by visualizing the attention weights. As shown in Fig. \ref{fig:example_frames}e, the attention of the heart rate model is commonly focused on the forehead, the earlobe and the carotid arteries. The earlobe has a large blood supply and the carotid arteries have the most significant pulse-induced motions. For BR measurement, the attention maps are more scattered, because respiration movement can transmit to any body part even including the hair. We also found high attention weights around the nose on many subjects, which suggests our CAN model uses subtle nasal flaring as a feature for respiratory tracking (see Supplemental Materials).

\section{Conclusions}
We have proposed the first end-to-end network for non-contact measurement of HR and BR. 
Our convolutional attention network allows spatial-temporal visualization of physiological distributions whilst learning color and motion information for recovering the physiological signals. We evaluated our method on three datasets of RGB videos and a dataset of IR videos. Our method outperformed all the prior state-of-the-art approaches that we compared against. The performance improvements were especially good for the tasks with increasing range and angular velocities of head rotation. We attribute this improvement to the end-to-end nature of the model which is able to learn an improved mapping between the video color and motion information. The participant dependent vs. independent performance as well as the transfer learning results shows that our supervised method does indeed generalize to other people, skin types and illumination conditions.

\printbibliography

\end{document}